\documentclass[letterpaper]{article} 
\usepackage{aaai2026}  
\usepackage{times}  
\usepackage{helvet}  
\usepackage{courier}  
\usepackage[hyphens]{url}  
\usepackage{graphicx} 
\usepackage{natbib}  
\usepackage{caption} 
\usepackage{algorithm}
\usepackage{algorithmic}
\usepackage{amssymb}
\usepackage{xcolor}
\usepackage{multirow}
\usepackage{booktabs}
\usepackage{amsmath} 

\usepackage{newfloat}
\usepackage{listings}
\DeclareCaptionStyle{ruled}{labelfont=normalfont,labelsep=colon,strut=off} 
\floatstyle{ruled}
\newfloat{listing}{tb}{lst}{}
\floatname{listing}{Listing}
\title{URaG: Unified Retrieval and Generation in Multimodal LLMs for Efficient Long Document Understanding}
\author{
    Yongxin Shi\textsuperscript{\rm 1}\equalcontrib,
    Jiapeng Wang\textsuperscript{\rm 1}\equalcontrib,
    Zeyu Shan\textsuperscript{\rm 1},
    Dezhi Peng\textsuperscript{\rm 2}\thanks{Corresponding author.},
    Zening Lin\textsuperscript{\rm 1},
    Lianwen Jin\textsuperscript{\rm 1 †},\\
}
\affiliations{
    \textsuperscript{\rm 1}South China University of Technology\\
    \textsuperscript{\rm 2}Huawei Technologies Co., Ltd.\\
    yongxin\_shi@foxmail.com, eelwjin@scut.edu.cn
}

\begin{document}

\maketitle


\begin{abstract}
Recent multimodal large language models (MLLMs) still struggle with long document understanding due to two fundamental challenges: information interference from abundant irrelevant content, and the quadratic computational cost of Transformer-based architectures. Existing approaches primarily fall into two categories: token compression, which sacrifices fine-grained details; and introducing external retrievers, which increase system complexity and prevent end-to-end optimization. To address these issues, we conduct an in-depth analysis and observe that MLLMs exhibit a human-like coarse-to-fine reasoning pattern: early Transformer layers attend broadly across the document, while deeper layers focus on relevant evidence pages. Motivated by this insight, we posit that the inherent evidence localization capabilities of MLLMs can be explicitly leveraged to perform retrieval during the reasoning process, facilitating efficient long document understanding. To this end, we propose \textbf{URaG}, a simple-yet-effective framework that \underline{U}nifies \underline{R}etrieval \underline{a}nd \underline{G}eneration within a single MLLM. URaG introduces a lightweight cross-modal retrieval module that converts the early Transformer layers into an efficient evidence selector, identifying and preserving the most relevant pages while discarding irrelevant content. This design enables the deeper layers to concentrate computational resources on pertinent information, improving both accuracy and efficiency. Extensive experiments demonstrate that URaG achieves state-of-the-art performance while reducing computational overhead by 44-56\%. The code is available at https://github.com/shi-yx/URaG.
\end{abstract}

\section{Introduction}
Document understanding plays a pivotal role in a wide range of real-world applications, such as information extraction, contract analysis, and report processing. 
While recent multimodal large language models (MLLMs) have shown impressive performance in processing single-page documents~\cite{zhu2025internvl3, bai2025qwen2}, the transition from single-page to multi-page document understanding introduces fundamental scalability and efficiency challenges that remain largely unresolved. 
The main challenges are twofold. 
First, the presence of a large volume of irrelevant content often leads to information interference.
Second, due to the quadratic computational complexity of Transformer-based architectures with respect to sequence length, processing excessively long sequences results in prohibitively high computational costs, significantly limiting the scalability of existing approaches.

\begin{figure}[t]
  \centering
  \includegraphics[width=\linewidth]{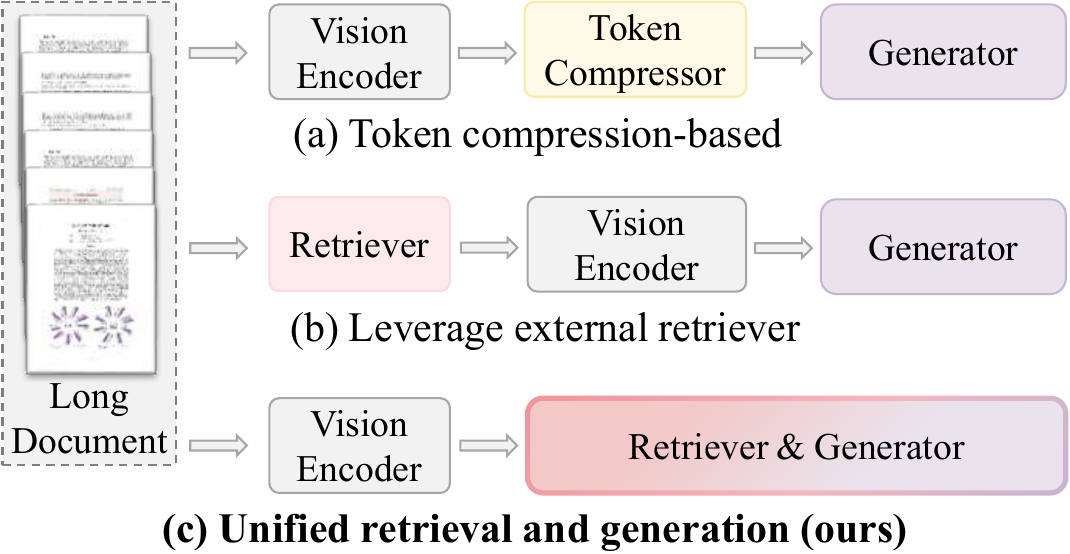}
  \caption{Comparison of different methods for long document understanding. Our method unifies retrieval and generation within a single MLLM, leveraging early-layer features for evidence retrieval during reasoning, which achieves efficient and accurate long document understanding.}
  \label{fig1}
\end{figure}

To address these challenges, 
existing MLLM-based approaches primarily follow two technical routes:
(1) The first route involves compressing the input tokens fed into the LLM. 
These methods~\cite{hu2024mplug, jia2024leopard} typically apply uniform compression to visual tokens across all pages of the document, thus reducing the token count and alleviating the computational burden.
However, this strategy inevitably sacrifices certain visual details during the compression process, potentially impairing the model’s capacity for fine-grained visual understanding.
(2) The second technical approach involves the introduction of an external retriever. 
These methods~\cite{zhang2024cream, chen2025lora} employ text-based or vision-based retrievers to extract the most relevant content from long documents, feeding only the retrieved subset into the MLLM. 
While effective in reducing computational overhead, it introduces high system complexity in real-world deployment, due to the reliance on a separate retrieval module. 
More importantly, it lacks end-to-end optimization: 
Since the retriever is typically trained independently from the MLLM, such systems are prone to suboptimal coordination and error propagation.

When processing long documents, humans rarely read every word carefully and sequentially. 
Instead, they adopt a coarse-to-fine comprehension strategy~\cite{leon2019selective, zou2023human}.
(1) Coarse-grained retrieval: Based on an initial understanding of the question (e.g., keywords, intent, and contextual cues), humans first leverage structural features of the document (e.g., layout, titles, and figures) to rapidly identify pages or regions that are likely to contain relevant information.
(2) Fine-grained reading: After locating candidate regions, humans engage in detailed reading to extract precise answers.
Motivated by this human reading behavior, we hypothesize that MLLMs exhibit a similar coarse-to-fine reasoning pattern when processing long documents. 
To verify this hypothesis, we conduct a systematic empirical study. 
Our analysis reveals that attention distribution evolves progressively across the Transformer layers of the LLM component: early layers tend to distribute attention uniformly across pages, whereas deeper layers increasingly concentrate attention on pages containing answer evidence. 
This shift in attention patterns provides compelling empirical support that MLLMs inherently perform human-like hierarchical reasoning for long document understanding.

Building on this insight, we posit that the inherent evidence localization capabilities of MLLMs can be explicitly leveraged to perform retrieval during the reasoning process, facilitating efficient long document understanding.
To this end, we propose a simple-yet-effective framework, \textbf{URaG}, which \underline{U}nifies  \underline{R}etrieval \underline{a}nd \underline{G}eneration within a single MLLM.
The key innovation of our framework lies in transforming the early layers (e.g., the first 6 layers) of the MLLM into an efficient retrieval system through a lightweight cross-modal retrieval module. 
This module consists of two linear layers with minimal additional parameters, processes the hidden states of the early layer to extract visual features and textual features from each page and the question, respectively. 
Using a contextualized late interaction mechanism~\cite{khattab2020colbert}, it computes relevance scores between text and vision to identify the top-k most relevant pages. 
These selected pages are preserved in the hidden states and propagated to the deeper layers for answer generation, while irrelevant content is discarded. 
By unifying retrieval and generation within a single model, our framework enables precise evidence localization, which effectively mitigates information interference and substantially reduces computational overhead.

The effectiveness of URaG is extensively verified on several commonly used benchmarks. 
Without bells and whistles, experimental results show that URaG achieves state-of-the-art performance. 
In addition, its computational efficiency is also empirically validated, further underscoring its practical advantages in real-world deployment.

In summary, our main contributions are as follows.
\begin{itemize}
    \item We present a systematic empirical study revealing that MLLMs inherently exhibit a human-like coarse-to-fine reasoning pattern when processing long documents.
    \item We propose URaG, an elegant framework that seamlessly integrates evidence retrieval and answer generation within a single MLLM, eliminating the need for external retrieval systems. Equipped with a lightweight cross-modal retrieval module, URaG explicitly leverages the inherent evidence localization capabilities of MLLMs to perform efficient and integrated retrieval.
    \item Extensive experiments demonstrate the effectiveness of our method, which enhances MLLMs’ long document comprehension capability while reducing computational overhead by 44-56\%.
\end{itemize}

\begin{figure*}[t]
\centering
\includegraphics[width=1\textwidth]{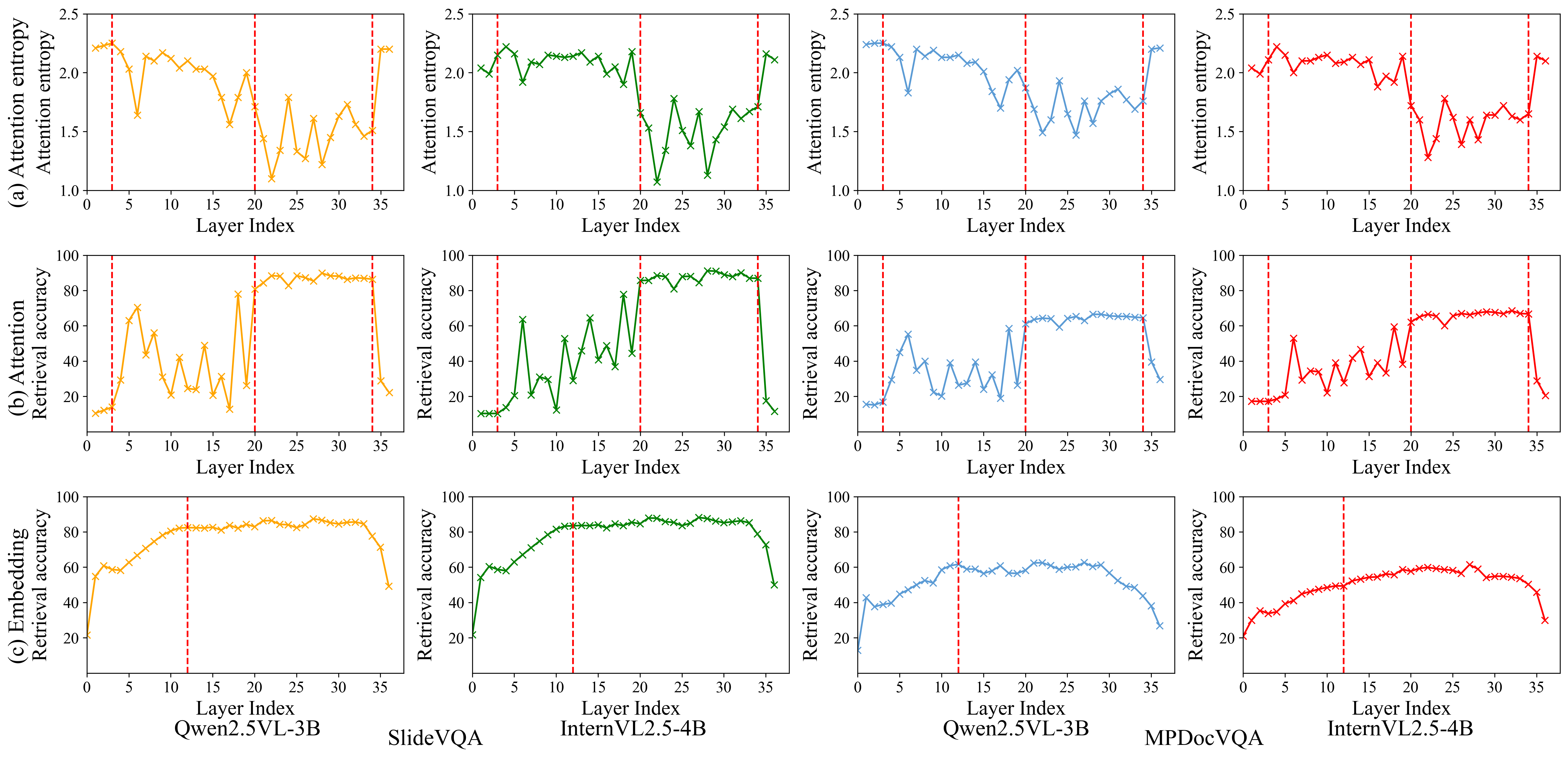} 
\caption{Analysis of MLLMs on long document understanding. (a) Attention entropy. (b) Attention-based retrieval accuracy. (c) Embedding-based retrieval accuracy.}
\label{fig3}
\end{figure*}

\section{Related Work}
\subsection{Long Document Understanding}
Existing methods for long document understanding can be primarily categorized into encoder-decoder-based and MLLM-based approaches.
\textbf{Encoder-decoder-based} methods are built upon encoder-decoder Transformer architectures, such as T5~\cite{raffel2020exploring}. Representative methods include:
Hi-VT5~\cite{tito2023hierarchical} summarizes key information from each page into special [PAGE] tokens for hierarchical decoding.
GRAM~\cite{blau2024gram} integrates local single-page encoding with global document-level layers, using learnable tokens and bias adaptation to enhance cross-page reasoning.
RM-T5~\cite{dong2024multi} employs recurrent memory to propagate information across pages sequentially.
\textbf{MLLM-based} methods primarily focus on optimizing the performance of multimodal large language models in long-sequence reasoning, with two main strategies:
(1) Compressing input tokens.
To alleviate computational costs, these methods focus on compressing the visual inputs before feeding them into the language model.
For instance, mPLUG-DocOwl2~\cite{hu2024mplug} introduces a high-resolution DocCompressor module that reduces each document image to 324 tokens.
Similarly, Leopard~\cite{jia2024leopard} proposes an adaptive high-resolution multi-image encoder, which dynamically allocates visual token sequences based on the resolution and aspect ratios of the input images.
(2) Incorporating external retrievers.
These approaches employ retrieval mechanisms to pre-select relevant content prior to MLLM inference.
For example, CREAM~\cite{zhang2024cream} adopts a coarse-to-fine retrieval pipeline that combines embedding-based similarity search, multi-round grouping, and LLM-based re-ranking to extract the most relevant text segments.
SV-RAG~\cite{chen2025lora} leverages the final hidden states of MLLMs for question-guided evidence retrieval, feeding only the selected content into the model for answer generation.
M3DocRAG~\cite{cho2024m3docrag} employs a multimodal retriever to identify relevant content prior to the MLLM.

\subsection{Document Retrieval}
Document retrieval approaches can be broadly categorized into text-based and vision-based methods.
\textbf{Text-Based retrieval} methods typically rely on Optical Character Recognition (OCR) to extract textual content from documents, followed by similarity computation between the extracted text and the query. 
These methods can be further categorized into sparse and dense retrieval techniques. Widely used sparse retrievers include:
TF-IDF~\cite{salton1983extended} calculates term relevance using word frequency and inverse document frequency;
BM25~\cite{robertson1995okapi} improves upon TF-IDF by introducing non-linear term frequency saturation and document length normalization, enhancing ranking robustness.
Dense retrieval methods encode text into continuous vector spaces, enabling semantic similarity matching.
DPR~\cite{karpukhin2020dense} uses a dual-encoder architecture to independently encode questions and passages.
SBERT~\cite{reimers2019sentence} produces sentence-level embeddings via a siamese BERT~\cite{devlin2019bert}.
BGE~\cite{xiao2024c} improves dense retrieval quality through self-knowledge distillation and careful curation of training data.
NV-Embed-v2~\cite{lee2024nv} introduces a latent attention-based pooling mechanism for aggregating token representations.
\textbf{Vision-based retrieval} methods directly encode document images, preserving both textual and layout information.
CLIP~\cite{radford2021learning} and SigLIP~\cite{zhai2023sigmoid} are commonly used to extract retrieval embeddings.
Some approaches further leverage MLLMs to jointly encode document images and textual queries for multimodal retrieval.
For instance, ColPali~\cite{faysse2024colpali} utilizes PaliGemma~\cite{beyer2024paligemma} to obtain token-level embeddings for each document page.
DSE~\cite{ma2024unifying} employs Phi-3-V~\cite{abdin2024phi} to encode each page into a single dense embedding, facilitating compact yet effective document representation.
MM-Embed~\cite{linmm} fine-tunes MLLM-based universal multimodal retrievers and prompts pre-trained MLLMs for zero-shot reranking over
retrieved candidates.

\begin{figure*}[t]
\centering
\includegraphics[width=0.9\textwidth]{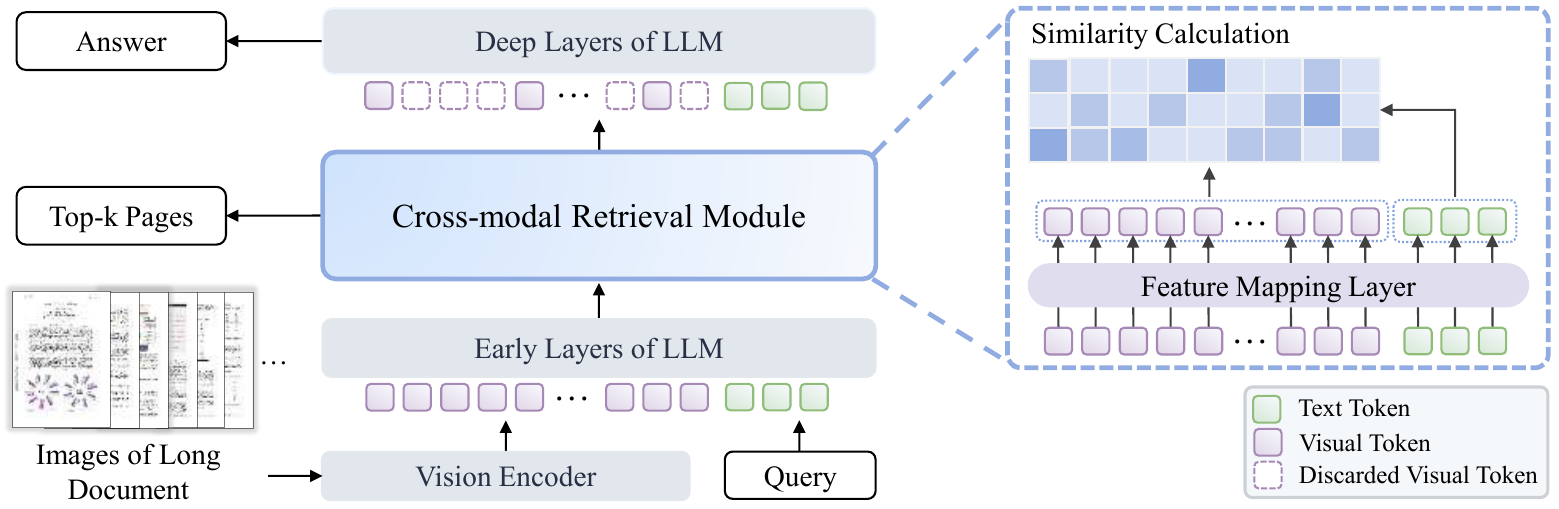} 
\caption{Overview of our URaG framework.}
\label{fig2}
\end{figure*}

\section{Analysis}
To investigate whether MLLMs exhibit a human-like coarse-to-fine reading behavior, we conduct a systematic empirical study on two representative MLLMs across the subsets of two long document understanding benchmarks. 
Specifically, we visualize the attention entropy across LLM layers, measuring how the generated tokens attend over input pages, as shown in Figure~\ref{fig3} (a).
We further evaluate the retrieval accuracy using attention weights as retrieval scores to identify evidence pages, as shown in Figure~\ref{fig3} (b). 
In addition, we compute query-to-visual similarity using hidden states from each layer as embeddings, enabling embedding-based retrieval, as shown in Figure~\ref{fig3} (c). 
From the visualizations and metrics, we observe the following trends:
(1) In the early layers (e.g., the first 3 layers), the attention entropy is high and attention-based retrieval accuracy is correspondingly low. 
This indicates that the model distributes attention relatively uniformly across all pages, reflecting a global and coarse-level reading stage.
(2) In the early-middle layers (e.g., layers 3–20), attention entropy exhibits a declining trend with fluctuations, while attention-based retrieval accuracy shows a corresponding upward trend. 
This suggests the model is progressively attempting to identify and focus on relevant evidence pages.
(3) In deeper layers (e.g., layers 20-34), the attention entropy remains low and attention-based retrieval accuracy stays consistently high, indicating a fine-grained reading stage where attention becomes highly concentrated on evidence pages. 
(4) In the final two layers, attention entropy increases again while attention-based retrieval accuracy slightly drops. This indicates the model revisits all input pages before the final answer, which is similar to how humans recheck the full document to ensure correctness.
In summary, these findings provide strong empirical evidence that MLLMs inherently follow a human-like coarse-to-fine reasoning pattern when processing long documents. 
This insight motivates us to explicitly harness their intrinsic evidence localization ability for unified retrieval-generation approaches.

Moreover, we find that embedding-based retrieval reaches consistently high accuracy (e.g., around layer 12) earlier than attention-based. 
This implies that semantic representations formed at mid-level layers are already sufficiently discriminative for evidence selection. In addition, embedding-based retrieval shows more stable performance across layers. These two observations motivate us to adopt embedding-based retrieval to develop our method.

\section{Methodology}
\subsection{Framework}
The proposed URaG is a unified method that integrates both retrieval and generation within a single model. 
As illustrated in Figure~\ref{fig2}, the framework consists of a multimodal large language model (MLLM) and a lightweight cross-modal retrieval module.
Specifically, given a long document composed of pages \(\{p_1, p_2, \ldots, p_n\}\), where \(n\) denotes the number of pages, and a user query \(Q\), each page image is processed by the vision encoder and the projector to obtain a sequence of visual tokens. 
The query \(Q\) is tokenized by a text tokenizer and then input into the LLM along with the visual tokens.
The retrieval module operates on the hidden states from an early layer (the sixth layer in our implementation) of the LLM to retrieve the top-\(k\) (\(k\) is set to 5 by default) pages most relevant to the query. 
The visual tokens corresponding to non-retrieved pages are directly discarded from the hidden states.
Subsequently, the deeper LLM layers attend only to the retained pages for answer generation.

\subsection{Cross-modal Retrieval Module}
The retrieval module is designed to identify the most relevant pages from multi-page inputs with respect to the query.
It is implemented with a lightweight structure consisting of two linear projections with GELU activation. 
Formally, given the early-layer hidden states \( H \in \mathbb{R}^{L \times D} \), the feature mapping layer is applied to reduce the dimensionality, yielding \( H' \in \mathbb{R}^{L \times D'} \), followed by L2 normalization to enhance feature consistency.
From \( H' \), the visual feature sequences for each document pages \( \{ E_v^{(1)}, E_v^{(2)}, \dots, E_v^{(n)} \} \) and the textual feature sequence of the query \( E_q \), are extracted based on positional indices.  
The similarity between the query text and each document page is computed using the widely adopted contextualized late interaction~\cite{khattab2020colbert}:

\begin{equation}
s_{q, v^{(p)}} = \sum_{i \in [|E_q|]} \max_{j \in [|E_v^{(p)}|]} E_{q_i} \cdot E_{v_j}^{(p) T} \text{.}
\end{equation}

Based on the similarity scores, the top-\(k\) pages are retained while others are discarded directly from hidden states, enabling subsequent layers to focus on relevant content, significantly reducing computational overhead.

\begin{table*}[t]
    \centering
    \resizebox{0.86\linewidth}{!}{
    \begin{tabular}{lccccccccc}
        \toprule
        \multirow{2}{*}{Method} & \multirow{2}{*}{\#Param} & \multicolumn{2}{c}{SlideVQA} & \multicolumn{2}{c}{MMLong} & \multicolumn{2}{c}{DUDE} & \multicolumn{2}{c}{MPDocVQA} \\
        \cmidrule(lr){3-4} \cmidrule(lr){5-6} \cmidrule(lr){7-8} \cmidrule(lr){9-10}
         & & Top1 & Top5 & Top1 & Top5 & Top1 & Top5 & Top1 & Top5 \\
        \midrule
        \multicolumn{9}{l}{Text-based} \\
        \midrule
        BM25~\cite{robertson1995okapi} & - & 69.3 & 91.1 & 25.3 & 47.6 & 58.4 & 89.8 & 59.7 & 87.8 \\
        SBERT~\cite{reimers2019sentence} & - & 73.0 & 91.0 & 44.7 & 70.2 & 61.7 & 90.2 & 67.8 & 93.3 \\
        BGE-M3~\cite{chen2024bge} & 568M & 74.3 & 92.0 & 42.7 & 66.6 & 60.1 & 90.1 & 66.8 & 92.7 \\
        BGE-large~\cite{xiao2024c} & 326M & 81.3 & 93.3 & 47.4 & 71.5 & 60.1 & 90.1 & 66.8 & 92.7 \\
        NV-Embed-v2~\cite{lee2024nv} & 7B & 82.2 & 94.3 & 47.4 & 69.0 & 68.8 & 93.9 & 74.3 & 95.2 \\
        \midrule
        \multicolumn{9}{l}{Vision-based} \\
        \midrule
        CLIP~\cite{radford2021learning} & 428M & 58.4 & 86.9 & 32.4 & 63.4 & 61.0 & 89.5 & 67.8 & 93.7 \\
        SigLIP~\cite{zhai2023sigmoid} & 878M & 66.2 & 90.1 & 44.9 & 69.4 & 59.4 & 89.7 & 64.9 & 91.9 \\
        ColPali~\cite{faysse2024colpali} & 3B & 90.2 & 98.2 & 60.3 & 80.2 & 68.5 & 93.3 & 73.6 & 95.6 \\
        MM-Embed~\cite{linmm} & 7B & 70.9 & 91.8 & 42.9 & 74.7 & 65.6 & 91.9 & 69.5 & 94.0 \\
        SV-RAG~\cite{chen2025lora} & 4B & 90.6 & 98.8 & \underline{64.8} & 84.8 & - & - & - & - \\
        \midrule
        URaG-3B (ours) & 3B & \underline{92.1} & \underline{98.9} & 63.0 & \underline{85.4} & \underline{83.0} & \textbf{97.0} & \underline{84.4} & \textbf{98.0} \\
        URaG-7B (ours) & 7B & \textbf{92.9} & \textbf{99.0} & \textbf{68.3} & \textbf{86.0} & \textbf{83.9} & \underline{96.9} & \textbf{84.5} & \textbf{98.0} \\
        \bottomrule
    \end{tabular}
    }
    \caption{Retrieval performance comparison with different methods. MMLong refers to MMLongBench-Doc.}
    \label{tab:retrieval}
\end{table*}

\subsection{Training strategy}
We adopt a two-stage training strategy to optimize our framework.
In the first stage, we pretrain the retrieval module to adapt it for the retrieval task. 
All model parameters are frozen except for those in the retrieval module, which is optimized using the retrieval loss~\cite{khattab2020colbert}:

\begin{equation}
\mathcal{L_\text{retrieval}} = \log\left( 1 + \exp(S_{\text{neg}} - S_{\text{pos}}) \right),
\end{equation}

where \(S_{\text{pos}}\) and \(S_{\text{neg}}\) represent the scores of positive and negative samples, respectively. They are calculated as follows.

\begin{equation}S_{\text{pos}} = \sum\limits_{i \in P} s_i\end{equation}

\begin{equation}
S_{\text{neg}} =
\begin{cases}
\sum\limits_{j \in N} s_j & \text{if } N < P \\
\sum\limits_{j \in \text{TopK}(\{s_k \mid k \in N \}, P)} s_j & \text{if } N \geq P
\end{cases}
\end{equation}
Here, \(P\) and \(N\) indicate the number of positive and negative samples, respectively, and \(s\) denotes the similarity score between the query and a document page.

In the second stage, the LoRA~\cite{hu2022lora} adapter is added to both the LLM and retrieval module, with other parameters kept frozen. The model is jointly optimized through the retrieval loss and generation loss.
\begin{equation}
\mathcal{L}_{\text{total}} = \mathcal{L}_{\text{retrieval}} + \mathcal{L}_{\text{generation}}, 
\end{equation}
where \(\mathcal{L}_{\text{generation}}\) is the cross-entropy loss for answer generation.
To facilitate adaptation to fine-grained visual features, we retain at most five pages after retrieval.
Ground-truth evidence pages are always kept to ensure information completeness, while remaining pages are selected by the highest retrieval scores.

\section{Experiments}
\subsection{Implementation Details}
Our model is available in two sizes: URaG-3B and URaG-7B, both built upon Qwen2.5-VL~\cite{bai2025qwen2}.
The retrieval module consists of two linear projection layers with GELU activation, which sequentially reduce the dimension of hidden states to 1024 and then to 512.
The model is trained with a batch size of 4 and 8 gradient accumulation steps with AdamW optimizer. 
During pretraining, the initial learning rate is set to $1 \times 10^{-4}$ with a warm-up ratio of 0.03, followed by a cosine decay schedule.
For fine-tuning, we adopt LoRA~\cite{hu2022lora} with a rank of 32, alpha of 64, and a dropout rate of 0.1. The loss weights of retrieval and generation are set to 1:1.
The warm-up and learning rate settings are the same as in pretraining.
The datasets for training including MPDocVQA~\cite{tito2023hierarchical}, DUDE~\cite{van2023document}, and SlideVQA~\cite{tanaka2023slidevqa}.
Both retrieval pretraining and joint fine-tuning are conducted for 1 epoch, respectively. 
Following prior work~\cite{xie2024wukong, chen2025lora}, we retain the top-5 pages in the retrieval module by default.
All experiments are conducted on 4 NVIDIA A6000 GPUs.

\subsection{Evaluation Metrics}
Following previous work~\cite{chen2025lora}, the evaluation involves retrieval and generation metrics.
The retrieval metrics include Top1 and Top5 accuracy.
The generation metrics include \textit{Average Normalized Levenshtein Similarity (ANLS)}~\cite{tito2023hierarchical} for MPDocVQA and DUDE, \textit{Exact Match (EM)}~\cite{tanaka2023slidevqa} for SlideVQA, \textit{Generalized Accuracy} and \textit{F1-score}~\cite{ma2024mmlongbench} for MMLongBench-Doc, and \textit{Generalized Accuracy score}~\cite{deng2024longdocurl} for LongDocURL.

\begin{table*}[!ht]
    \centering
    \resizebox{0.85\linewidth}{!}{
    \begin{tabular}{lcccccc}
        \toprule
        Method & \#Param & MPDocVQA & DUDE & SlideVQA & LongDocURL \\
        \midrule
        LayoutLMv3~\cite{huang2022layoutlmv3} & 125M & 55.1 & 20.3 & - & - \\
        Hi-VT5~\cite{tito2023hierarchical} & 316M & 61.8 & 35.7 & - & - \\
        DocFormerv2~\cite{appalaraju2024docformerv2} & 784M & 76.4 & 48.4 & - & - \\
        GRAM~\cite{blau2024gram} & 859M & 83.0 & 53.4 & - & - \\
        Llama-3.2~\cite{meta_llama32_2024} & 11B & 57.6 & 20.8 & - & 9.2 \\ 
        LLaVA-Next-Interleave~\cite{li2024llava2} & 7B & 39.9 & 24.0 & - & 14.1 \\
        Idefics3~\cite{laurenccon2023obelics} & 8B & 67.2 & 38.7 & 39.9 & - \\
        mPLUG-DocOwl2~\cite{hu2024mplug} & 8B & 69.4 & 46.7 & 24.6 & 5.3 \\
        CREAM~\cite{zhang2024cream} & 14B & 65.3 & 52.5 & - & - \\
        Qwen2-VL~\cite{wang2024qwen2} & 7B & 82.1 & 45.9 & 59.9 & 30.6 \\ 
        InternVL2.5~\cite{chen2024expanding} & 4B & 74.9 & 40.7 & 45.2 & 24.1 \\
        InternVL3~\cite{zhu2025internvl3} & 8B & 80.8 & 47.4 & 54.4 & 38.7 \\
        PDF-WuKong~\cite{xie2024wukong} & 8.5B & 76.9 & \underline{56.1} & - & - \\
        Qwen2.5-VL~\cite{bai2025qwen2} & 3B & 84.4 & 50.6 & 59.1 & 40.0 \\
        Qwen2.5-VL~\cite{bai2025qwen2} & 7B & \underline{87.2} & 55.0 & \underline{66.4} & \underline{51.1} \\
        \midrule
        URaG-3B (ours) & 3B & 86.0 & 54.1 & 63.8 & 41.5 \\
        URaG-7B (ours) & 7B & \textbf{88.2} & \textbf{57.6} & \textbf{72.1} & \textbf{52.2} \\
        \bottomrule
    \end{tabular}
    }
    \caption{Performance comparison with different methods on MPDocVQA, DUDE, SlideVQA, and LongDocURL benchmarks.}
    \label{tab:mpdoc}

    \vspace{1em} 

    \resizebox{\linewidth}{!}{
    \begin{tabular}{lccccccccccc}
        \toprule
        \multirow{2}{*}{Method} & \multirow{2}{*}{\#Param} & \multicolumn{5}{c}{Evidence Modalities} & \multicolumn{3}{c}{Evidence Locations} & \multicolumn{2}{c}{Overall} \\
        \cmidrule(lr){3-7} \cmidrule(lr){8-10} \cmidrule(lr){11-12}
        & & TXT & LAY & CHA & TAB & IMG & SIN & MUL & UNA & ACC & F1 \\
        \midrule
        DeepSeek-VL~\cite{lu2024deepseek} & 7B & 7.2 & 6.5 & 1.6 & 5.2 & 7.6 & 5.2 & 7.0 & 12.8 & 7.4 & 5.4 \\
        Idefics2~\cite{laurenccon2024matters} & 8B & 9.0 & 10.6 & 4.8 & 4.1 & 8.7 & 7.7 & 7.2 & 5.0 & 7.0 & 6.8 \\
        MiniCPM-V2.5~\cite{yao2024minicpm} & 8B & 11.9 & 10.8 & 5.1 & 5.9 & 12.2 & 9.5 & 9.5 & 4.5 & 8.5 & 8.6 \\
        InternLM-XC2-4KHD~\cite{dong2024internlm} & 8B & 9.9 & 14.3 & 7.7 & 6.3 & 13.0 & 12.6 & 7.6 & 9.6 & 10.3 & 9.8 \\
        mPLUG-DocOwl 1.5~\cite{hu2024mplug1} & 8B & 8.2 & 8.4 & 2.0 & 3.4 & 9.9 & 7.4 & 6.4 & 6.2 & 6.9 & 6.3 \\
        Qwen-VL\cite{bai2023qwen} & 10B & 5.5 & 9.0 & 5.4 & 2.2 & 6.9 & 5.2 & 7.1 & 6.2 & 6.1 & 5.4 \\
        Monkey~\cite{li2024monkey} & 10B & 6.8 & 7.2 & 3.6 & 6.7 & 9.4 & 6.6 & 6.2 & 6.2 & 6.2 & 5.6 \\
        CogVLM2-LLaMA3~\cite{hong2024cogvlm2} & 19B & 3.7 & 2.7 & 6.0 & 3.2 & 6.9 & 3.9 & 5.3 & 3.7 & 4.4 & 4.0 \\
        InternVL2.5~\cite{chen2024internvl} & 4B & 20.4 & 15.1 & 8.9 & 12.5 & 16.6 & 19.7 & 12.4 & 13.5 & 15.9 & 15.6 \\
        InternVL3~\cite{zhu2025internvl3} & 8B & 28.4 & 26.7 & 13.1 & 20.7 & 24.9 & 29.0 & 16.3 & 26.2 & 24.1 & 23.1 \\
        SV-RAG~\cite{chen2025lora} & 4B & 26.3 & 22.1 & \underline{25.0} & 20.7 & 25.2 & 34.0 & 10.6 & 15.7 & 23.0 & 24.2 \\
        M3DocRAG~\cite{cho2024m3docrag} & 10B & \underline{30.0} & 23.5 & 18.9 & 20.1 & 20.8 & 32.4 & 14.8 & 5.8 & 21.0 & 22.6 \\
        Qwen2.5-VL~\cite{bai2025qwen2} & 3B & 29.0 & 26.8 & 18.6 & 17.4 & 22.4 & 31.8 & 15.7 & 27.8 & 25.5 & 24.1 \\
        Qwen2.5-VL~\cite{bai2025qwen2} & 7B & \underline{30.0} & \underline{26.9} & 22.4 & 20.6 & \underline{22.9} & 33.4 & \textbf{17.5} & 24.7 & 26.2 & 25.1 \\
        \midrule
        URaG-3B (ours) & 3B & 27.9 & 24.2 & 23.4 & \underline{25.2} & 21.4 & \underline{36.9} & 13.5 & \textbf{48.9} & \underline{31.1} & \underline{28.7} \\
        URaG-7B (ours) & 7B & \textbf{33.6} & \textbf{27.7} & \textbf{29.3} & \textbf{27.5} & \textbf{27.2} & \textbf{42.7} & \underline{16.9} & \underline{43.5} & \textbf{33.8} & \textbf{32.8} \\
        \bottomrule
    \end{tabular}
    }
    \caption{Performance comparison with different methods on MMLongBench-Doc. Generalized accuracy across 5 evidence sources: pure text (TXT), layout (LAY), charts (CHA), tables (TAB), and images (IMG). Results are further categorized by the number of evidence pages: single-page (SIN), cross-page (MUL), and unanswerable (UNA) questions.}
    \label{tab:mmlong}
\end{table*}

\subsection{Evidence Page Retrieval}
We evaluate the evidence retrieval performance of URaG on MPDocVQA~\cite{tito2023hierarchical}, DUDE~\cite{van2023document}, SlideVQA~\cite{tanaka2023slidevqa}, and MMLongBench-Doc~\cite{ma2024mmlongbench}, and compare it with both the text-based and vision-based retrievers. 
For text-based methods, we employ Paddle-OCR~\cite{cui2025paddleocr30technicalreport} to extract textual content from document images for retrieval, except on MMLongBench-Doc, where the PDF parser is used.
As shown in Table~\ref{tab:retrieval}, our URaG consistently outperforms all compared methods across datasets.
This highlights the effectiveness of our integrated retrieval mechanism, which explicitly leverages the inherent capabilities of MLLMs via a lightweight retrieval module, without requiring complex designs or extensive training.

\subsection{Main Results}
We evaluate the performance of URaG on various long document understanding benchmarks, including MPDocVQA, DUDE, SlideVQA, LongDocURL, and MMLongBench-Doc. The results are demonstrated in Table~\ref{tab:mpdoc} and~\ref{tab:mmlong}, respectively. 
Based on the results, we draw the following key conclusions.  
Firstly, our method demonstrates strong effectiveness in long document understanding, achieving state-of-the-art performance across multiple benchmarks. 
Notably, our method significantly outperforms previous methods on SlideVQA and MMLongBench-Doc, which contain substantially long inputs with average lengths of 20 and 47.5 pages, respectively. 
This highlights the robustness and scalability of URaG in handling extended document contexts.
Secondly, from the perspective of evidence types, URaG shows superior performance on single-page questions, indicating that evidence localized within a single page is more accurately retrieved.
This can also be attributed to the distribution of training data, where questions with single-page evidence are more prevalent. 
Thirdly, URaG performs particularly well on visually intensive question types such as charts (CHA) and images (IMG). 
This indicates the effectiveness of the cross-modal retrieval module in performing semantic matching between textual and visual content.

\subsection{Comparison with Baseline Method}
To ensure a fair comparison, we fine-tune the baseline model~\cite{bai2025qwen2} using the same training data and settings as URaG. 
As shown in Table~\ref{tab:baseline}, our method significantly outperforms the baseline, demonstrating its effectiveness.
Notably, even without any fine-tuning of the MLLM backbone, by simply inserting a pretrained retrieval module, URaG already surpasses the fully fine-tuned baseline.
This highlights the strength of our cross-modal retrieval module, which effectively exploits the MLLM’s inherent evidence localization ability to perform retrieval.
With this simple and lightweight design, URaG achieves strong performance without large-scale training.
Moreover, on the LongDocURL dataset, fine-tuning leads to degraded performance. 
We attribute this to potential overfitting or domain mismatch between training and evaluation data. 
In contrast, URaG without any fine-tuning maintains robust performance, further demonstrating its generalization ability and practical applicability.

\begin{table}[t]
\centering
\resizebox{1\linewidth}{!}{
\begin{tabular}{lccc}
\toprule
Method & SlideVQA & MMLong & LongDoc \\
\midrule
Baseline & 59.1 & 25.5 & 40.0 \\
Baseline w/ SFT & 61.9 & 29.1 & 37.3 \\
URaG-3B w/o finetune & \underline{62.1} & \underline{29.4} & \textbf{43.1} \\
URaG-3B & \textbf{63.8} & \textbf{31.1} & \underline{41.5} \\
\bottomrule
\end{tabular}
}
\caption{Comparison with baseline method. MMLong and LongDoc refer to MMLongBench-Doc and LongDocURL, respectively.}
\label{tab:baseline}
\end{table}

\subsection{Ablation Study}
\subsubsection{The Position of Retrieval Module.} 
We investigate the impact of inserting the cross-modal retrieval module at different layers of the LLM using the 3B-size model.
As shown in Table~\ref{tab:layer}, the results reveal that inserting the retrieval module at early to mid layers (e.g., layer 6) leads to the best overall performance.
This can be attributed to two main factors.
First, although the top1 retrieval accuracy continues to improve slightly at deeper layers (e.g., layer 12 or 18), the top5 accuracy already reaches saturation at layer 6, and moving the retrieval module deeper yields only marginal gains.
Second, placing the retrieval module at earlier layers allows the subsequent LLM layers to concentrate on reasoning over a reduced set of visual features, effectively filtering out irrelevant information and improving the final answer generation quality.
In summary, these results support our design choice: inserting the retrieval module at early layers captures sufficiently discriminative semantic representations for evidence selection, while leaving deeper layers to focus on reasoning.

\begin{table}[h]
\centering
\resizebox{0.93\linewidth}{!}{
\begin{tabular}{ccccccc}
\toprule
\multirow{2}{*}{Layer} & \multicolumn{4}{c}{SlideVQA} & \multicolumn{2}{c}{MMLong} \\ 
\cmidrule(lr){2-5} \cmidrule(lr){6-7}
\multicolumn{1}{c}{} & Top1 & Top5 & EM$^*$ & EM & Top5 & GACC \\
\midrule
2 & 89.0 & 98.4 & 67.4 & 63.7 & 82.0 & \underline{31.0} \\
6 & 92.1 & 98.9 & \textbf{68.5} & \textbf{63.8} & \underline{85.4} & \textbf{31.1} \\
12 & \underline{93.1} & \textbf{99.2} & \underline{67.1} & \underline{62.9} & \textbf{85.6} & \underline{31.0} \\
18 & \textbf{93.5} & \textbf{99.2} & \underline{67.1} & 62.3 & \underline{85.4} & 30.6 \\
\bottomrule
\end{tabular}
}
\caption{Ablation study of the cross-modal retrieval module insertion at different layers within the LLM. $^*$ indicates metrics evaluated on the single-page subset of SlideVQA while retaining only the top-1 page after the retrieval module.}
\label{tab:layer}
\end{table}

\subsubsection{Two-stage Training Strategy.}
We conduct an ablation study to evaluate the effectiveness of our two-stage training strategy. 
As shown in Table~\ref{tab:two_stage}, the results demonstrate that both the retrieval pretraining stage and the joint fine-tuning stage contribute to the final performance.
Concretely, the retrieval capability of the model is limited without pretraining, as the retrieval module is trained from random initialization. 
Moreover, fine-tuning on the pretrained model further improves both retrieval accuracy and overall understanding.

\begin{table}[h]
\centering
\resizebox{0.88\linewidth}{!}{
\begin{tabular}{cccccc}
\toprule
\multicolumn{1}{l}{\multirow{2}{*}{Pretrain}} & \multicolumn{1}{c}{\multirow{2}{*}{Fintune}} & \multicolumn{2}{c}{SlideVQA} & \multicolumn{2}{c}{MMLong} \\
\cmidrule(lr){3-4} \cmidrule(lr){5-6}
\multicolumn{1}{l}{} & \multicolumn{1}{c}{} & Top5 & EM & Top5 & GACC \\
\midrule
$\times$ & \checkmark & 98.5 & 62.6 & 82.0 & 30.7 \\
\checkmark & $\times$ & 98.8 & 62.1 & 84.0 & 29.4 \\
\checkmark & \checkmark & \textbf{98.9} & \textbf{63.8} & \textbf{85.4} & \textbf{31.1} \\
\bottomrule
\end{tabular}
}
\caption{Ablation study of our two-stage training strategy. MMLong refers to MMLongBench-Doc.}
\label{tab:two_stage}
\end{table}

\subsection{Computational Efficiency}
To evaluate the computational efficiency of URaG, we conduct experiments on the SlideVQA dataset, where each question is originally associated with 20 document pages. 
To simulate longer input sequences, we duplicate the pages for each question. 
We compare URaG with two representative approaches: the token-compression-based mPLUG-DocOwl2~\cite{hu2024mplug} and the external-retriever-based SV-RAG~\cite{chen2025lora}, by integrating their techniques into a common baseline.
Model computational cost is measured in floating-point operations (FLOPs).
As shown in Table~\ref{tab:efficency}, URaG achieves consistently higher computational efficiency on longer inputs compared with other methods. 
When the number of input pages increases from 20 to 100, URaG reduces FLOPs by 44.0\% to 55.8\% compared to the baseline. 
These results demonstrate the effectiveness of our method in reducing computational complexity for long document inputs.
In addition, we analyze the number of parameters introduced by the cross-modal retrieval module and find that it accounts for only 0.05\% to 0.07\% of the total model parameters, which is nearly negligible. 

\begin{table}[!h]
    \centering
    \resizebox{1\linewidth}{!}{
        \begin{tabular}{lccc}
        \toprule
        \multirow{2}{*}{Method} & \multicolumn{3}{c}{Pages} \\
        \cmidrule(lr){2-4}
        & 20 & 60 & 100 \\
        \midrule
        Baseline & 415.9 & 1246.4 & 2076.9 \\
        mPLUG-DocOwl2~\shortcite{hu2024mplug} & \textbf{208.3} & \underline{624.5} & \underline{1040.7} \\
        SV-RAG~\shortcite{chen2025lora} & 393.1 & 969.5 & 1546.0 \\
        \midrule
        URaG-7B (ours) & \underline{232.8} & \textbf{574.9} & \textbf{917.0} \\
        Reduction & \textcolor{green}{-44.0\%} & \textcolor{green}{-53.9\%} & \textcolor{green}{-55.8\%} \\
        \bottomrule
    \end{tabular}
    }
    \caption{Computational efficiency comparison with different methods in terms of FLOPs (T). }
    \label{tab:efficency}
\end{table}

\section{Conclusion}
In this paper, we propose URaG, a unified framework that unifies retrieval and generation within a single multimodal large language model (MLLM) for efficient long document understanding.  
URaG introduces a lightweight cross-modal retrieval module that explicitly leverages the inherent evidence localization capabilities of early layers of MLLMs, enabling it to identify and retain only the most relevant pages during the reasoning process.
Extensive experiments across multiple long document understanding benchmarks demonstrate that URaG not only achieves state-of-the-art performance but also significantly reduces computational overhead by 44-56\%. 
We believe this work not only provides a practical solution but also valuable insights into a novel paradigm for long document understanding.

\section{Acknowledgments}
This research is supported in part by the National Natural Science Foundation of China (Grant No.:62476093) and Huawei-SCUT Research Project Fund (No. TC20250611036).

\bibliography{aaai2026}

@String{Springer = "Springer-Verlag" }

@article{raffel2020exploring,
  title={Exploring the limits of transfer learning with a unified text-to-text {Transformer}},
  author={Raffel, Colin and Shazeer, Noam and Roberts, Adam and Lee, Katherine and Narang, Sharan and Matena, Michael and Zhou, Yanqi and Li, Wei and Liu, Peter J},
  journal={Journal of machine learning research},
  volume={21},
  number={140},
  pages={1--67},
  year={2020}
}

@article{tito2023hierarchical,
  title={Hierarchical multimodal {Transformers} for multipage {DocVQA}},
  author={Tito, Rub{\`e}n and Karatzas, Dimosthenis and Valveny, Ernest},
  journal={Pattern Recognition},
  volume={144},
  pages={109834},
  year={2023},
  publisher={Elsevier}
}

@inproceedings{blau2024gram,
  title={{GRAM}: Global reasoning for multi-page {VQA}},
  author={Blau, Tsachi and Fogel, Sharon and Ronen, Roi and Golts, Alona and Ganz, Roy and Ben Avraham, Elad and Aberdam, Aviad and Tsiper, Shahar and Litman, Ron},
  booktitle={Proc. CVPR},
  pages={15598--15607},
  year={2024}
}

@inproceedings{dong2024multi,
  title={Multi-page document {VQA} with recurrent memory {Transformer}},
  author={Dong, Qi and Kang, Lei and Karatzas, Dimosthenis},
  booktitle={Proc. ICDAR Workshop},
  pages={57--70},
  year={2024},
  organization={Springer}
}

@article{hu2024mplug,
  title={{mPLUG-DocOwl2}: High-resolution compressing for {OCR}-free multi-page document understanding},
  author={Hu, Anwen and Xu, Haiyang and Zhang, Liang and Ye, Jiabo and Yan, Ming and Zhang, Ji and Jin, Qin and Huang, Fei and Zhou, Jingren},
  journal={arXiv preprint arXiv:2409.03420},
  year={2024}
}

@article{jia2024leopard,
  title={Leopard: A vision language model for text-rich multi-image tasks},
  author={Jia, Mengzhao and Yu, Wenhao and Ma, Kaixin and Fang, Tianqing and Zhang, Zhihan and Ouyang, Siru and Zhang, Hongming and Jiang, Meng and Yu, Dong},
  journal={arXiv preprint arXiv:2410.01744},
  year={2024}
}

@inproceedings{zhang2024cream,
  title={{CREAM}: {Coarse-to-fine} retrieval and multi-modal efficient tuning for document {VQA}},
  author={Zhang, Jinxu and Yu, Yongqi and Zhang, Yu},
  booktitle={Proc. ACM MM},
  pages={925--934},
  year={2024}
}

@article{xie2024wukong,
  title={{WuKong}: A large multimodal model for efficient long {PDF} reading with end-to-end sparse sampling},
  author={Xie, Xudong and Yan, Hao and Yin, Liang and Liu, Yang and Ding, Jing and Liao, Minghui and Liu, Yuliang and Chen, Wei and Bai, Xiang},
  journal={arXiv preprint arXiv:2410.05970},
  year={2024}
}

@article{salton1983extended,
  title={Extended boolean information retrieval},
  author={Salton, Gerard and Fox, Edward A and Wu, Harry},
  journal={Communications of the ACM},
  volume={26},
  number={11},
  pages={1022--1036},
  year={1983},
  publisher={ACM New York, NY, USA}
}

@article{robertson1995okapi,
  title={Okapi at {TREC-3}},
  author={Robertson, Stephen E and Walker, Steve and Jones, Susan and Hancock-Beaulieu, Micheline M and Gatford, Mike and others},
  journal={Nist Special Publication Sp},
  volume={109},
  pages={109},
  year={1995},
  publisher={National Instiute of Standards \& Technology}
}

@inproceedings{karpukhin2020dense,
  title={Dense passage retrieval for open-domain question answering.},
  author={Karpukhin, Vladimir and Oguz, Barlas and Min, Sewon and Lewis, Patrick SH and Wu, Ledell and Edunov, Sergey and Chen, Danqi and Yih, Wen-tau},
  booktitle={Proc. EMNLP},
  pages={6769--6781},
  year={2020}
}

@inproceedings{reimers2019sentence,
  title={{Sentence-BERT}: Sentence embeddings using siamese {BERT}-networks},
  author={Reimers, Nils and Gurevych, Iryna},
  booktitle={Proc. EMNLP},
  pages={3982--3992},
  year={2019}
}

@inproceedings{devlin2019bert,
  title={{BERT}: Pre-training of deep bidirectional {Transformers} for language understanding},
  author={Devlin, Jacob and Chang, Ming-Wei and Lee, Kenton and Toutanova, Kristina},
  booktitle={Proc. NAACL},
  pages={4171--4186},
  year={2019}
}

@inproceedings{xiao2024c,
  title={{C-Pack}: Packed resources for general {Chinese} embeddings},
  author={Xiao, Shitao and Liu, Zheng and Zhang, Peitian and Muennighoff, Niklas and Lian, Defu and Nie, Jian-Yun},
  booktitle={Proc. SIGIR},
  pages={641--649},
  year={2024}
}

@article{lee2024nv,
title={Nv-Embed: Improved techniques for training {LLMs} as generalist embedding models},
  author={Lee, Chankyu and Roy, Rajarshi and Xu, Mengyao and Raiman, Jonathan and Shoeybi, Mohammad and Catanzaro, Bryan and Ping, Wei},
  journal={arXiv preprint arXiv:2405.17428},
  year={2024}
}

@article{ma2024unifying,
  title={Unifying multimodal retrieval via document screenshot embedding},
  author={Ma, Xueguang and Lin, Sheng-Chieh and Li, Minghan and Chen, Wenhu and Lin, Jimmy},
  journal={arXiv preprint arXiv:2406.11251},
  year={2024}
}

@article{beyer2024paligemma,
  title={{PaliGemma}: A versatile {3B} {VLM} for transfer},
  author={Beyer, Lucas and Steiner, Andreas and Pinto, Andr{\'e} Susano and Kolesnikov, Alexander and Wang, Xiao and Salz, Daniel and Neumann, Maxim and Alabdulmohsin, Ibrahim and Tschannen, Michael and Bugliarello, Emanuele and others},
  journal={arXiv preprint arXiv:2407.07726},
  year={2024}
}

@article{abdin2024phi,
  title={Phi-3 technical report: A highly capable language model locally on your phone},
  author={Abdin, Marah and Aneja, Jyoti and Awadalla, Hany and Awadallah, Ahmed and Awan, Ammar Ahmad and Bach, Nguyen and Bahree, Amit and Bakhtiari, Arash and Bao, Jianmin and Behl, Harkirat and others},
  journal={arXiv preprint arXiv:2404.14219},
  year={2024}
}

@inproceedings{van2023document,
  title={Document understanding dataset and evaluation ({DUDE})},
  author={Van Landeghem, Jordy and Tito, Rub{\`e}n and Borchmann, {\L}ukasz and Pietruszka, Micha{\l} and Joziak, Pawel and Powalski, Rafal and Jurkiewicz, Dawid and Coustaty, Micka{\"e}l and Anckaert, Bertrand and Valveny, Ernest and others},
  booktitle={Proc. ICCV},
  pages={19528--19540},
  year={2023}
}

@inproceedings{tanaka2023slidevqa,
  title={{SlideVQA}: A dataset for document visual question answering on multiple images},
  author={Tanaka, Ryota and Nishida, Kyosuke and Nishida, Kosuke and Hasegawa, Taku and Saito, Itsumi and Saito, Kuniko},
  booktitle={Proc. AAAI},
  volume={37},
  pages={13636--13645},
  year={2023}
}

@inproceedings{radford2021learning,
  title={Learning transferable visual models from natural language supervision},
  author={Radford, Alec and Kim, Jong Wook and Hallacy, Chris and Ramesh, Aditya and Goh, Gabriel and Agarwal, Sandhini and Sastry, Girish and Askell, Amanda and Mishkin, Pamela and Clark, Jack and others},
  booktitle={Proc. ICML},
  pages={8748--8763},
  year={2021},
  organization={PmLR}
}

@inproceedings{zhai2023sigmoid,
  title={Sigmoid loss for language image pre-training},
  author={Zhai, Xiaohua and Mustafa, Basil and Kolesnikov, Alexander and Beyer, Lucas},
  booktitle={Proc. ICCV},
  pages={11975--11986},
  year={2023}
}

@inproceedings{chen2025lora,
  title={{SV-RAG}: {LoRA}-contextualizing adaptation of large multimodal models for multi-page document Understanding},
  author={Chen, Jian and Zhang, Ruiyi and Zhou, Yufan and Yu, Tong and Dernoncourt, Franck and Gu, Jiuxiang and Rossi, Ryan A and Chen, Changyou and Sun, Tong},
  booktitle={Proc. ICLR},
  year={2025}
}

@inproceedings{faysse2024colpali,
  title={Colpali: Efficient document retrieval with vision language models},
  author={Faysse, Manuel and Sibille, Hugues and Wu, Tony and Omrani, Bilel and Viaud, Gautier and Hudelot, C{\'e}line and Colombo, Pierre},
  booktitle={Proc. ICLR},
  year={2025}
}

@inproceedings{huang2022layoutlmv3,
  title={{LayoutLMv3}: Pre-training for document ai with unified text and image masking},
  author={Huang, Yupan and Lv, Tengchao and Cui, Lei and Lu, Yutong and Wei, Furu},
  booktitle={Proc. ACM MM},
  pages={4083--4091},
  year={2022}
}

@inproceedings{appalaraju2024docformerv2,
  title={{DocFormerv2}: Local features for document understanding},
  author={Appalaraju, Srikar and Tang, Peng and Dong, Qi and Sankaran, Nishant and Zhou, Yichu and Manmatha, R},
  booktitle={Proc. AAAI},
  volume={38},
  pages={709--718},
  year={2024}
}

@article{laurenccon2023obelics,
  title={{OBELICS}: An open web-scale filtered dataset of interleaved image-text documents},
  author={Lauren{\c{c}}on, Hugo and Saulnier, Lucile and Tronchon, L{\'e}o and Bekman, Stas and Singh, Amanpreet and Lozhkov, Anton and Wang, Thomas and Karamcheti, Siddharth and Rush, Alexander and Kiela, Douwe and others},
  journal={Proc. NeurIPS},
  volume={36},
  pages={71683--71702},
  year={2023}
}

@article{bai2025qwen2,
  title={{Qwen2.5-VL} technical report},
  author={Bai, Shuai and Chen, Keqin and Liu, Xuejing and Wang, Jialin and Ge, Wenbin and Song, Sibo and Dang, Kai and Wang, Peng and Wang, Shijie and Tang, Jun and others},
  journal={arXiv preprint arXiv:2502.13923},
  year={2025}
}

@article{hu2024mplug1,
  title={{mPLUG-DocOwl 1.5}: Unified structure learning for ocr-free document understanding},
  author={Hu, Anwen and Xu, Haiyang and Ye, Jiabo and Yan, Ming and Zhang, Liang and Zhang, Bo and Li, Chen and Zhang, Ji and Jin, Qin and Huang, Fei and others},
  journal={arXiv preprint arXiv:2403.12895},
  year={2024}
}

@article{wang2024qwen2,
  title={{Qwen2-VL}: Enhancing vision-language model's perception of the world at any resolution},
  author={Wang, Peng and Bai, Shuai and Tan, Sinan and Wang, Shijie and Fan, Zhihao and Bai, Jinze and Chen, Keqin and Liu, Xuejing and Wang, Jialin and Ge, Wenbin and others},
  journal={arXiv preprint arXiv:2409.12191},
  year={2024}
}

@article{li2024llava2,
  title={{LLaVA-NeXT-Interleave}: Tackling multi-image, video, and {3D} in large multimodal models},
  author={Li, Feng and Zhang, Renrui and Zhang, Hao and Zhang, Yuanhan and Li, Bo and Li, Wei and Ma, Zejun and Li, Chunyuan},
  journal={arXiv preprint arXiv:2407.07895},
  year={2024}
}

@misc{meta_llama32_2024,
  author       = {{Meta AI}},
  title        = {{LLaMA} 3.2: Revolutionizing Edge AI and Vision with Open, Customizable Models},
  howpublished = {\url{https://ai.meta.com/blog/llama-3-2-connect-2024-vision-edge-mobile-devices/}},
  year         = {2024}
}

@article{lu2024deepseek,
  title={{DeepSeek-VL}: towards real-world vision-language understanding},
  author={Lu, Haoyu and Liu, Wen and Zhang, Bo and Wang, Bingxuan and Dong, Kai and Liu, Bo and Sun, Jingxiang and Ren, Tongzheng and Li, Zhuoshu and Yang, Hao and others},
  journal={arXiv preprint arXiv:2403.05525},
  year={2024}
}

@article{laurenccon2024matters,
  title={What matters when building vision-language models?},
  author={Lauren{\c{c}}on, Hugo and Tronchon, L{\'e}o and Cord, Matthieu and Sanh, Victor},
  journal={Proc. NeurIPS},
  volume={37},
  pages={87874--87907},
  year={2024}
}

@article{bai2023qwen,
  title={Qwen technical report},
  author={Bai, Jinze and Bai, Shuai and Chu, Yunfei and Cui, Zeyu and Dang, Kai and Deng, Xiaodong and Fan, Yang and Ge, Wenbin and Han, Yu and Huang, Fei and others},
  journal={arXiv preprint arXiv:2309.16609},
  year={2023}
}

@inproceedings{li2024monkey,
  title={Monkey: Image resolution and text label are important things for large multi-modal models},
  author={Li, Zhang and Yang, Biao and Liu, Qiang and Ma, Zhiyin and Zhang, Shuo and Yang, Jingxu and Sun, Yabo and Liu, Yuliang and Bai, Xiang},
  booktitle={Proc. CVPR},
  pages={26763--26773},
  year={2024}
}

@article{hong2024cogvlm2,
  title={{CogVLM2}: Visual language models for image and video understanding},
  author={Hong, Wenyi and Wang, Weihan and Ding, Ming and Yu, Wenmeng and Lv, Qingsong and Wang, Yan and Cheng, Yean and Huang, Shiyu and Ji, Junhui and Xue, Zhao and others},
  journal={arXiv preprint arXiv:2408.16500},
  year={2024}
}

@inproceedings{chen2024internvl,
  title={{InternVL}: Scaling up vision foundation models and aligning for generic visual-linguistic tasks},
  author={Chen, Zhe and Wu, Jiannan and Wang, Wenhai and Su, Weijie and Chen, Guo and Xing, Sen and Zhong, Muyan and Zhang, Qinglong and Zhu, Xizhou and Lu, Lewei and others},
  booktitle={Proc. CVPR},
  pages={24185--24198},
  year={2024}
}

@article{chen2024expanding,
  title={Expanding performance boundaries of open-source multimodal models with model, data, and test-time scaling},
  author={Chen, Zhe and Wang, Weiyun and Cao, Yue and Liu, Yangzhou and Gao, Zhangwei and Cui, Erfei and Zhu, Jinguo and Ye, Shenglong and Tian, Hao and Liu, Zhaoyang and others},
  journal={arXiv preprint arXiv:2412.05271},
  year={2024}
}

@article{cho2024m3docrag,
  title={{M3DocRAG}: Multi-modal retrieval is what you need for multi-page multi-document understanding},
  author={Cho, Jaemin and Mahata, Debanjan and Irsoy, Ozan and He, Yujie and Bansal, Mohit},
  journal={arXiv preprint arXiv:2411.04952},
  year={2024}
}

@article{yao2024minicpm,
  title={{MiniCPM-V}: A {GPT-4V} level {MLLM} on your phone},
  author={Yao, Yuan and Yu, Tianyu and Zhang, Ao and Wang, Chongyi and Cui, Junbo and Zhu, Hongji and Cai, Tianchi and Li, Haoyu and Zhao, Weilin and He, Zhihui and others},
  journal={arXiv preprint arXiv:2408.01800},
  year={2024}
}

@article{dong2024internlm,
  title={{InternLM-XComposer2-4KHD}: A pioneering large vision-language model handling resolutions from 336 pixels to {4K HD}},
  author={Dong, Xiaoyi and Zhang, Pan and Zang, Yuhang and Cao, Yuhang and Wang, Bin and Ouyang, Linke and Zhang, Songyang and Duan, Haodong and Zhang, Wenwei and Li, Yining and others},
  journal={Proc. NeurIPS},
  volume={37},
  pages={42566--42592},
  year={2024}
}

@inproceedings{khattab2020colbert,
  title={{ColBERT}: Efficient and effective passage search via contextualized late interaction over {BERT}},
  author={Khattab, Omar and Zaharia, Matei},
  booktitle={Proc. SIGIR},
  pages={39--48},
  year={2020}
}

@article{ma2024mmlongbench,
  title={{MMLongBench-Doc}: Benchmarking long-context document understanding with visualizations},
  author={Ma, Yubo and Zang, Yuhang and Chen, Liangyu and Chen, Meiqi and Jiao, Yizhu and Li, Xinze and Lu, Xinyuan and Liu, Ziyu and Ma, Yan and Dong, Xiaoyi and others},
  journal={Proc. NeurIPS},
  volume={37},
  pages={95963--96010},
  year={2024}
}

@article{deng2024longdocurl,
  title={{LongDocURL}: a comprehensive multimodal long document benchmark integrating understanding, reasoning, and locating},
  author={Deng, Chao and Yuan, Jiale and Bu, Pi and Wang, Peijie and Li, Zhong-Zhi and Xu, Jian and Li, Xiao-Hui and Gao, Yuan and Song, Jun and Zheng, Bo and others},
  journal={arXiv preprint arXiv:2412.18424},
  year={2024}
}

@misc{cui2025paddleocr30technicalreport,
      title={{PaddleOCR} 3.0 Technical Report}, 
      author={Cheng Cui and Ting Sun and Manhui Lin and Tingquan Gao and Yubo Zhang and Jiaxuan Liu and Xueqing Wang and Zelun Zhang and Changda Zhou and Hongen Liu and Yue Zhang and Wenyu Lv and Kui Huang and Yichao Zhang and Jing Zhang and Jun Zhang and Yi Liu and Dianhai Yu and Yanjun Ma},
      year={2025},
      eprint={2507.05595},
      archivePrefix={arXiv},
      primaryClass={cs.CV},
      url={https://arxiv.org/abs/2507.05595}, 
}

@inproceedings{hu2022lora,
  title={{LoRA}: Low-rank adaptation of large language models.},
  author={Hu, Edward J and Shen, Yelong and Wallis, Phillip and Allen-Zhu, Zeyuan and Li, Yuanzhi and Wang, Shean and Wang, Lu and Chen, Weizhu and others},
  booktitle={Proc. ICLR},
  year={2022}
}

@article{chen2024bge,
title={BGE M3-Embedding: Multi-lingual, multi-functionality, multi-granularity text embeddings through self-knowledge distillation},
  author={Chen, Jianlv and Xiao, Shitao and Zhang, Peitian and Luo, Kun and Lian, Defu and Liu, Zheng},
  journal={arXiv preprint arXiv:2402.03216},
  year={2024}
}

@article{zou2023human,
  title={Human attention during goal-directed reading comprehension relies on task optimization},
  author={Zou, Jiajie and Zhang, Yuran and Li, Jialu and Tian, Xing and Ding, Nai},
  journal={Elife},
  volume={12},
  pages={RP87197},
  year={2023},
  publisher={eLife Sciences Publications Limited}
}

@article{leon2019selective,
  title={Selective attention to question-relevant text information precedes high-quality summaries: Evidence from eye movements},
  author={Le{\'o}n, Jos{\'e} A and Moreno, Jos{\'e} David and Escudero, Inmaculada and Kaakinen, Johanna K},
  journal={Journal of Eye Movement Research},
  volume={12},
  number={1},
  pages={10--16910},
  year={2019}
}

@article{zhu2025internvl3,
  title={{InternVL3}: Exploring advanced training and test-time recipes for open-source multimodal models},
  author={Zhu, Jinguo and Wang, Weiyun and Chen, Zhe and Liu, Zhaoyang and Ye, Shenglong and Gu, Lixin and Tian, Hao and Duan, Yuchen and Su, Weijie and Shao, Jie and others},
  journal={arXiv preprint arXiv:2504.10479},
  year={2025}
}

@inproceedings{linmm,
  title={{MM-Embed}: Universal Multimodal Retrieval with Multimodal LLMs}, 
  author={Lin, Sheng-Chieh and Lee, Chankyu and Shoeybi, Mohammad and Lin, Jimmy and Catanzaro, Bryan and Ping, Wei},
  booktitle={Proc. ICLR},
  year={2025}
}

\newpage
\appendix
\section*{Appendix} 

\subsection{Details of Analysis}
\begin{figure}[h]
\centering
\includegraphics[width=\linewidth]{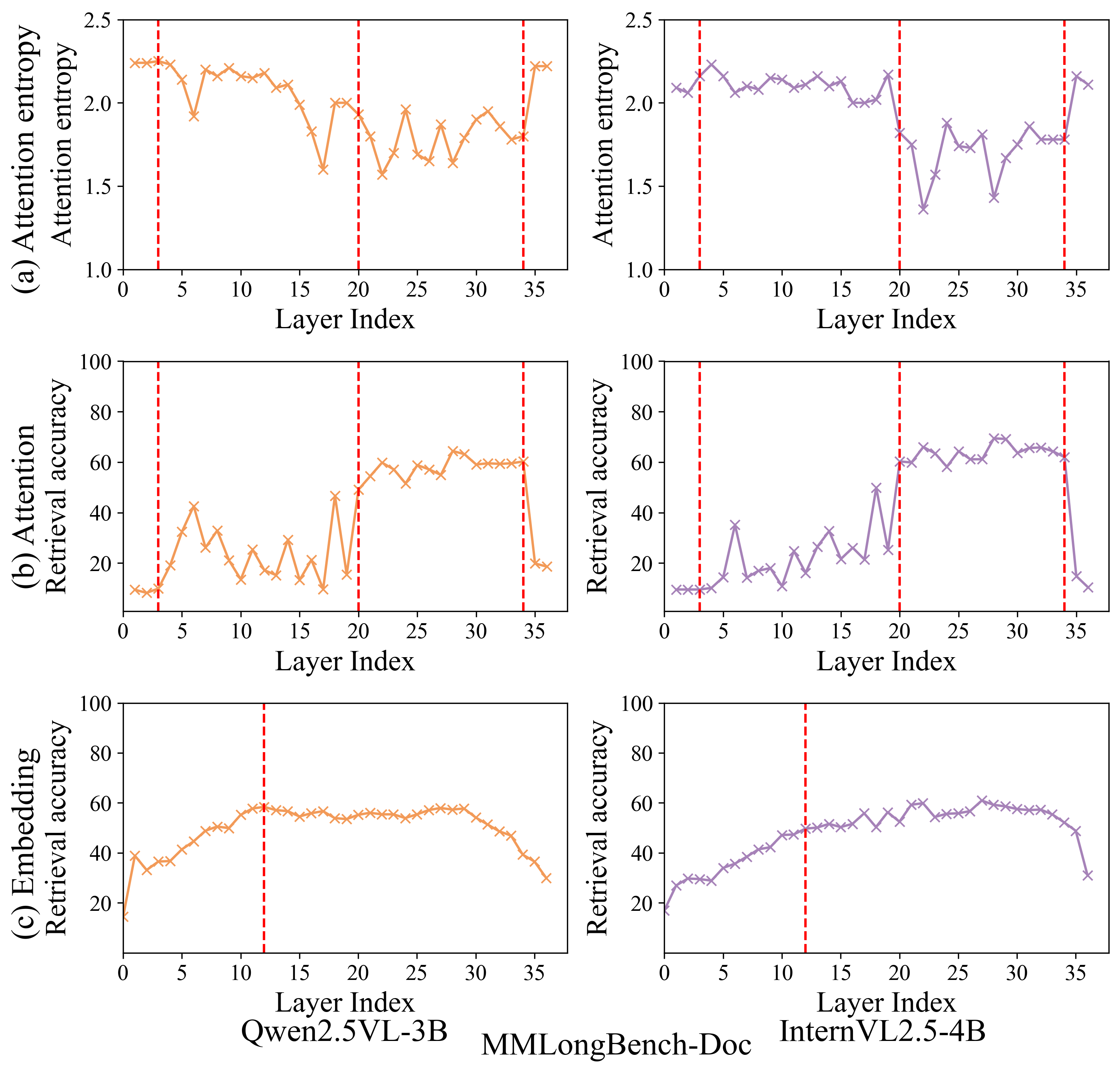} 
\caption{Analysis of MLLMs on long document understanding. (a) Attention entropy. (b) Attention-based retrieval accuracy. (c) Embedding-based retrieval accuracy. }
\label{fig4}
\end{figure}

Due to the constraints of computational resources, for samples with more than 10 pages, we crop them to a continuous span of 10 pages. 
Additionally, we adjust the configurations to control the resolution of the input image: the ``max\_pixels'' is set to 200,704 for Qwen2.5-VL~\cite{bai2025qwen2}, and the ``max\_num'' is set to 1 for InternVL2.5~\cite{chen2024expanding}.
The attention entropy is computed as follows:

\begin{align}
p_i &= \frac{w_i}{\sum_j w_j}, \\
entropy &= -\sum_i p_i \log(p_i),
\end{align}

where \(w_i\) denotes the attention weight assigned to page \(i\).
Specifically, it is computed as follows.
\begin{equation}
w_i = \sum_{h=1}^{H} \sum_{v \in \text{Page}_i} A^{(l,h)}_{g, v}
\end{equation}
where $A^{(l,h)}_{g, t}$ denotes the attention score from the generated token $g$ to token $v$ at layer $l$ and head $h$, 
$H$ is the total number of attention heads, and $\text{Page}_i$ denotes the set of token indices belonging to page $i$.
The embedding similarity is computed following Equation (1) in the Methodology section of the main paper.
We further report the results on the MMLongBench-Doc~\cite{ma2024mmlongbench} benchmark, as shown in Figure~\ref{fig4}, which provides further empirical support for the analysis discussed in the main paper.

\subsection{Details of Computational Efficiency}
We compare URaG with two representative approaches by integrating their techniques into a common baseline: the token-compression-based mPLUG-DocOwl2~\cite{hu2024mplug} and the external-retriever-based SV-RAG~\cite{chen2025lora}.
Specifically, for mPLUG-DocOwl2, we incorporate the high-resolution DocCompressor module after the vision encoder of Qwen2.5VL~\cite{bai2025qwen2} to reduce the sequence length of each document image to 324 tokens, and compute the overall FLOPs.
For SV-RAG, we follow its original design by adding two separate sets of LoRA~\cite{hu2022lora} adapters to Qwen2.5VL to construct the retriever and generator, respectively.
All document pages are processed by the retriever, and the retrieved top-5 pages are sent to the generator. 
The total computational cost is computed by summing the FLOPs of both modules.
In addition, we report results for the 3B-size variant of each method.
As shown in Table~\ref{tab:efficency2}, URaG-3B consistently achieves higher computational efficiency under longer inputs. 
When the number of input pages increases from 20 to 100, URaG-3B reduces FLOPs by 34.8\% to 44.0\% compared to the baseline.
These results further highlight the computational efficiency of URaG.

\begin{table}[!h]
    \centering
    \resizebox{1\linewidth}{!}{
        \begin{tabular}{lccc}
        \toprule
        \multirow{2}{*}{Method} & \multicolumn{3}{c}{Pages} \\
        \cmidrule(lr){2-4}
        & 20 & 60 & 100 \\
        \midrule
        Baseline & 243.5 & 730.0 & 1215.2 \\
        mPLUG-DocOwl2~\shortcite{hu2024mplug} & \textbf{152.4} & \underline{457.0} & \underline{761.6} \\
        SV-RAG~\shortcite{chen2025lora} & 230.3 & 568.5 & 906.6 \\
        \midrule
        URaG-3B (ours) & \underline{158.9} & \textbf{419.5} & \textbf{680.2} \\
        Reduction & \textcolor{green}{-34.8\%} & \textcolor{green}{-42.5\%} & \textcolor{green}{-44.0\%} \\
        \bottomrule
    \end{tabular}
    }
    \caption{Computational efficiency comparison with different methods in terms of FLOPs (T). }
    \label{tab:efficency2}
\end{table}

\begin{table}[!h]
\centering
{
\begin{tabular}{lccc}
\toprule
\multirow{2}{*}{Method} & \multicolumn{2}{c}{\#Params} & \multirow{2}{*}{Proportion (\%)} \\
\cmidrule(lr){2-3}
 & Retrieval Module & Total &  \\
\midrule
URaG-3B & 2.5M & 3.5B & 0.07 \\
URaG-7B & 4.0M & 7.7B & 0.05 \\
\bottomrule
\end{tabular}}
\caption{Proportion of the cross-modal retrieval module parameters in total model size.}
\label{tab:ratio}
\end{table}

\subsection{Parameter Efficiency}
As shown in Table~\ref{tab:ratio}, we analyze the number of parameters introduced by the cross-modal retrieval module and find that it accounts for only 0.05\% to 0.07\% of the total model parameters, which is nearly negligible. 

\subsection{Additional Implementation Details}
To limit the sequence length of visual tokens, we set the hyperparameters ``min\_pixels'' and ``max\_pixels'' as 100,352 and 802,816, respectively.
During inference, we set the temperature to 1.0 and top-p to 1.0.
The random seed is fixed to 42 across the experiments to ensure reproducibility.

\subsection{Additional Experiences of Computational Efficiency}
We record the average inference time per question and the peak GPU memory usage. As shown in Table~\ref{tab:Additional_comparison}, when the number of input pages increases from 20 to 100, our method achieves a time reduction of 16.7\% to 41.6\% and a memory reduction of 31.1\% to 51.3\% compared to the baseline. These results demonstrate the effectiveness of our method in accelerating inference and reducing memory consumption.

\begin{table}[!h]
    \centering
    \resizebox{1\linewidth}{!}{
    \begin{tabular}{lccc}
        \toprule
        Method & Page & Time(s) & Mem(GB) \\
        \midrule
        Qwen2.5VL & 20 & 3.66 & 14.30 \\
        URaG(ours) & 20 & 3.05~\textcolor{green}{(-16.67\%)} & 9.85~\textcolor{green}{(-31.12\%)} \\
        \midrule
        Qwen2.5VL & 40 & 8.67 & 21.40 \\
        URaG(ours) & 40 & 6.11~\textcolor{green}{(-29.53\%)} & 12.61~\textcolor{green}{(-41.07\%)} \\
        \midrule
        Qwen2.5VL & 60 & 15.62 & 28.51 \\
        URaG(ours) & 60 & 9.80~\textcolor{green}{(-37.26\%)} & 15.34~\textcolor{green}{(-46.19\%)} \\
        \midrule
        Qwen2.5VL & 80 & 24.35 & 35.62 \\
        URaG(ours) & 80 & 13.82~\textcolor{green}{(-43.24\%)} & 18.07~\textcolor{green}{(-49.27\%)} \\
        \midrule
        Qwen2.5VL & 100 & 32.07 & 42.73 \\
        URaG(ours) & 100 & 18.74~\textcolor{green}{(-41.57\%)} & 20.81~\textcolor{green}{(-51.30\%)} \\
        \bottomrule
    \end{tabular}}
    \caption{Additional Efficiency comparison of URaG and baseline model.}
    \label{tab:Additional_comparison}
\end{table}

\subsection{Additional Experiments}
To validate the generalizability of our proposed framework, we implement URaG based on InternVL2.5-4B~\cite{chen2024expanding}, and train it using the same data and settings described in the main paper.
The results are shown in Table~\ref{tab:baseline}, our method outperforms the baseline across multiple benchmarks, demonstrating the effectiveness and versatility of the proposed framework.

\begin{table}[h]
\centering
\begin{tabular}{lccc}
\toprule
Method & SlideVQA & MMLong & LongDocURL \\
\midrule
Baseline & 45.2 & 15.9 & 24.0 \\
URaG-4B & \textbf{51.9} & \textbf{16.8} & \textbf{29.5} \\
\bottomrule
\end{tabular}
\caption{Comparison URaG-4B with baseline method. MMLong refers to MMLongBench-Doc.}
\label{tab:baseline}
\end{table}

\subsection{Visualization of Retrieval}
\begin{figure}[t]
\centering
\includegraphics[width=0.99\linewidth]{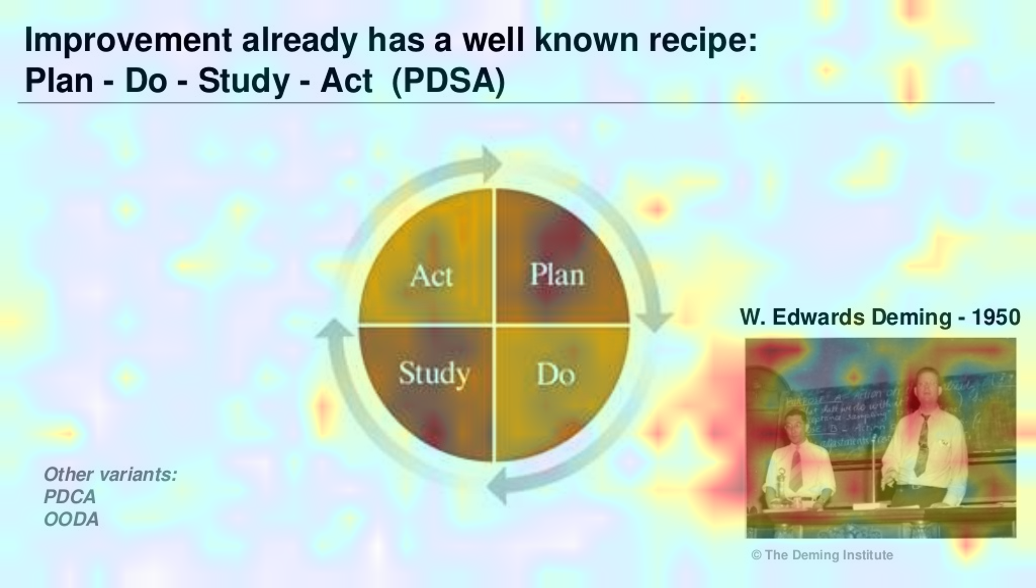} 
\caption{Similarity heatmap for the query: ``How many people are standing in front of the chalkboard in the photograph?''. }
\label{v1}
\end{figure}

\begin{figure}[ht]
\centering
\includegraphics[width=0.99\linewidth]{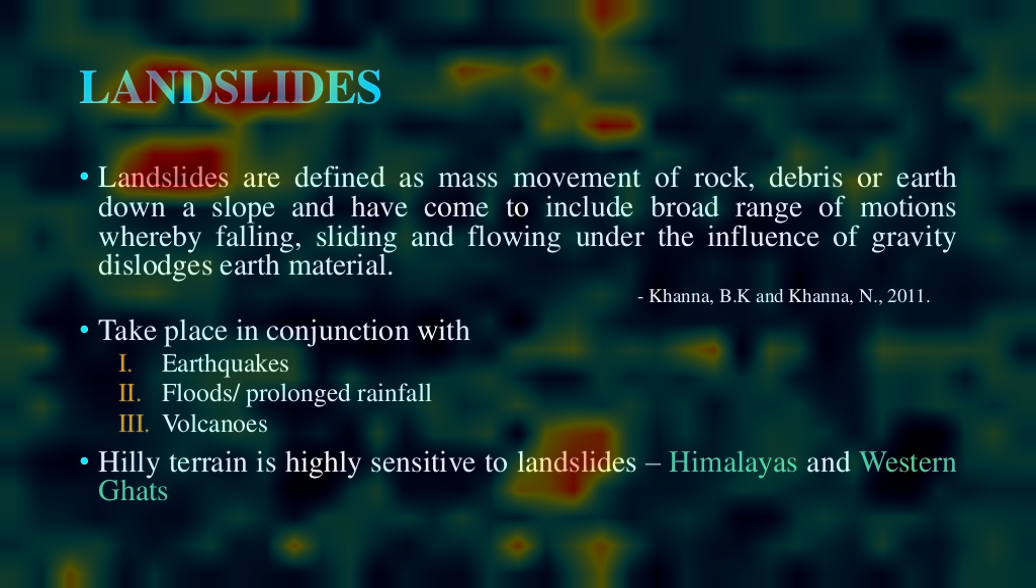} 
\caption{Similarity heatmap for the query: ``What are landslides?''. }
\label{v2}
\end{figure}

To better understand the retrieval behavior of our method, we visualize the patch-level embedding similarity between the query and document pages using URaG-7B. 
As shown in Figure~\ref{v1} and~\ref{v2}, the model effectively identifies relevant regions corresponding to the query.
For instance, for the question ``How many people are standing in front of the chalkboard in the photograph?'', the model highlights visual regions associated with ``people'' and the ``chalkboard''.
For the question ``What are landslides?'', the model accurately locates the keyword “landslides”.
These results indicate that the hidden states of the MLLMs encode fine-grained semantic alignment between the query and the visual content during reasoning. 
Our retrieval module effectively exploits these capabilities to perform evidence retrieval.

\subsection{Qualitative Results}
\begin{figure*}[ht]
\centering
\includegraphics[width=0.85\textwidth]{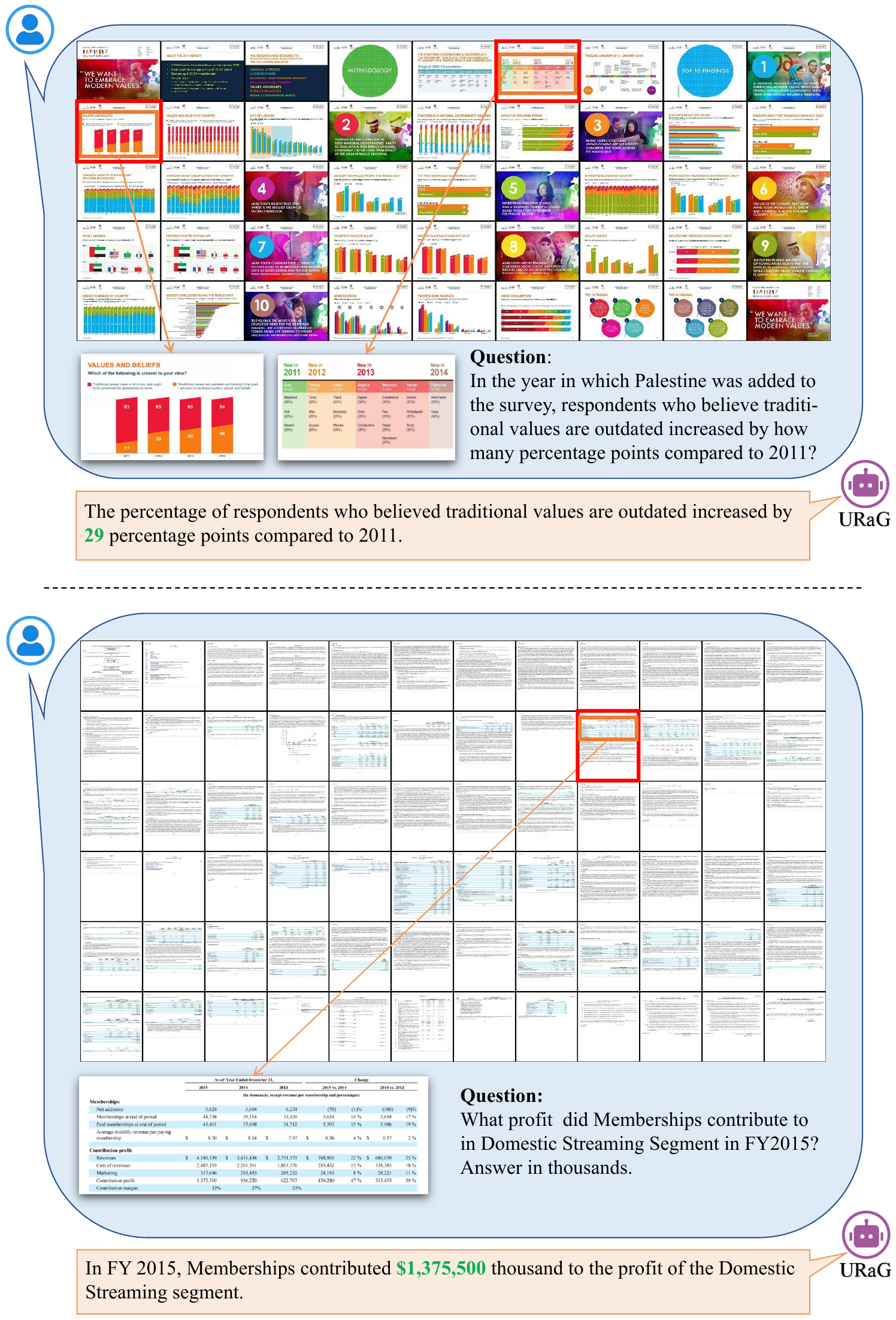} 
\caption{Qualitative results of our URaG.}
\label{case}
\end{figure*}

As illustrated in Figure~\ref{case}, URaG effectively extracts key information from long documents spanning dozens of pages, whether they are visually rich or text-dense.

\subsection{Limitation}
URaG employ a fixed top-$k$ retrieval strategy, where $k$ is set to 5 by default.
In cases where crucial evidence is dispersed across more than $k$ pages or concentrated within fewer pages, such a rigid selection could either omit essential information or introduce redundant inputs, thereby affecting both retrieval completeness and generation efficiency.
A potential direction for future work is to design a dynamic or adaptive retrieval mechanism that adjusts $k$ based on query difficulty or similarity confidence.

\end{document}